\documentclass[10pt,twocolumn,letterpaper]{article}

\usepackage[pagenumbers]{cvpr} 

\definecolor{cvprblue}{rgb}{0.21,0.49,0.74}
\usepackage[pagebackref,breaklinks,colorlinks,allcolors=cvprblue]{hyperref}

\def\paperID{11318} 
\def\confName{CVPR}
\def\confYear{2025}

\usepackage{graphicx}
\usepackage{booktabs}
\usepackage{makecell}
\usepackage{multicol}
\usepackage{multirow}

\usepackage{eurosym}
\usepackage{comment}
\usepackage{bbding,pifont}
\usepackage{soul}
\usepackage{tabularx}
\usepackage{pifont}
\usepackage{subcaption}
\usepackage{wrapfig}

\usepackage{amsmath}
\usepackage{amssymb}
\usepackage{booktabs}
\usepackage{colortbl}

\usepackage[accsupp]{axessibility}  

\newcommand{\cmark}{\ding{51}}%
\newcommand{\xmark}{\ding{55}}

\newcommand{\GB}{GCM}
\newcommand{\globalmodule}{global context module}
\newcommand{\globalbranch}{global context branch}

\newcommand{\capfirstglobalmodule}{Global context module}

\newcommand{\LB}{PIM}
\newcommand{\localmodule}{proximity integration module}
\newcommand{\localbranch}{proximity integration branch}

\newcommand{\capfirstlocalmodule}{Proximity integration module}

\newcommand{\baselineOne}{Pre-cropping}
\newcommand{\baselineTwo}{Post-cropping}

\newcommand{\roi}{ROI}

\newcommand{\context}{context}
\newcommand{\Context}{Context}
\newcommand{\task}{LocalSR}
\newcommand{\methodfull}{context-based local super-resolution}
\newcommand{\method}{CLSR}
\definecolor{bgcolor}{rgb}{0.9,0.9,0.9}

\title{LocalSR: Image Super-Resolution in Local Region}

\author{Bo Ji \qquad Angela Yao \\
National University of Singapore\\
{\tt\small \{jibo,ayao\}@comp.nus.edu.sg}
}

\begin{document}
\maketitle

\begin{abstract}
Standard single-image super-resolution (SR) upsamples and restores entire images. Yet several real-world applications require higher resolutions only in specific regions, such as license plates or faces, making the super-resolution of the entire image, along with the associated memory and computational cost, unnecessary. We propose a novel task, called LocalSR, to restore only local regions of the low-resolution image.  For this problem setting, we propose a context-based local super-resolution (CLSR) to super-resolve only specified regions of interest (\textit{ROI}) while leveraging the entire image as \textit{context}. Our method uses three parallel processing modules: a base module for super-resolving the ROI, a global context module for gathering helpful features from across the image, and a proximity integration module for concentrating on areas surrounding the ROI, progressively propagating features from distant pixels to the target region. Experimental results indicate that our approach, with its reduced low complexity, outperforms variants that focus exclusively on the ROI.
\end{abstract}

\section{Introduction}

\label{sec:intro}
Standard image super-resolution recovers high-resolution (HR) images from the low-resolution (LR) counterpart. 
For state-of-the-art methods~\cite{lim2017enhanced,zhang2018image,liang2021swinir,zhang2022perception}, the typical inference process takes the entire LR image as input and outputs the entire corresponding HR image.  
For some applications, it is unnecessary to super-resolve the entire image, especially if there are compute or data constraints. For example, for surveillance settings, faces and license plates are more relevant; for close-up photo-editing, only the magnified portion must be displayed. 
Beyond such applications, the GPU or device memory also sets an upper bound to the image size a model can super-resolve. 

\begin{figure*}[t!]
    \centering
    \includegraphics[width=0.8\linewidth]{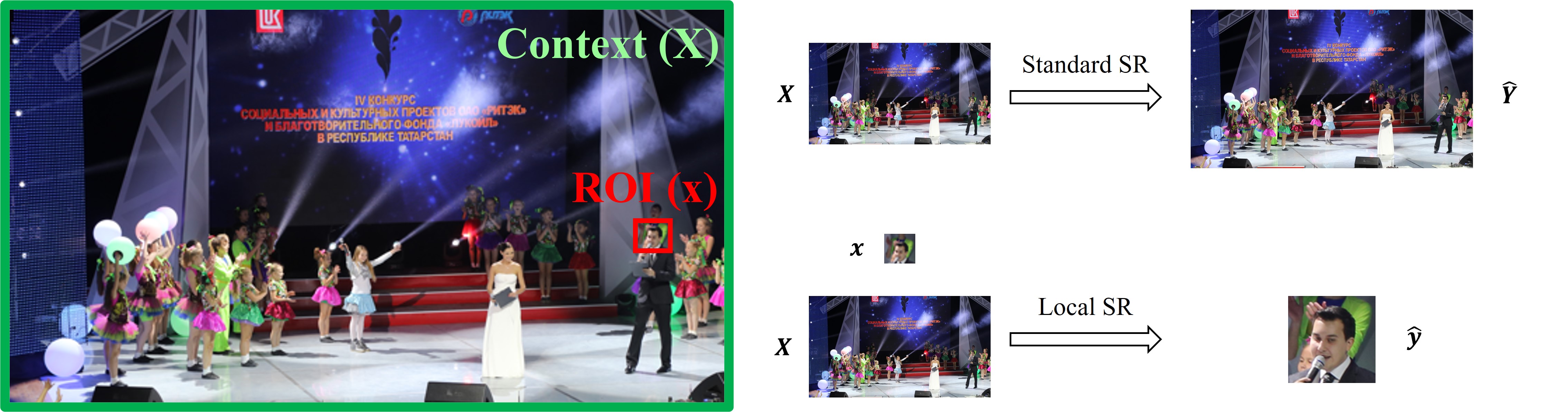}
    \caption{\textbf{Standard SR vs. Local SR}. Local SR focuses on enhancing specific region (\eg face) with less emphasis on other areas.}
    \label{fig:local_sr}
\end{figure*}

In response to these requirements, we introduce a novel task called local super-resolution (\task{}). 
\task{} concentrates on the high-quality restoration of a designated local region within an image rather than the entire image. Throughout this paper, we refer to the region targeted for restoration as the Region of Interest (\textit{\roi{}}) and the original LR image as the \textit{\context{}}, which includes the \roi{} and provides restoration information. The context does not require detailed processing and can be recovered simply, through \eg bi-linear interpolation.
Figure~\ref{fig:local_sr} illustrates an example of the application and shows the difference between standard SR and our proposed \task{}. 
The objective may be to visualize a person's face in a photo, disregarding the background or other individuals. In this context, it is worthwhile to investigate how the entire original image can be used to improve the quality of the local region of interest.

There are two straightforward approaches to \task{}. The first, which we refer to as a `pre-cropping', crops the \roi{} from the LR image for restoration; the second, which we call `post-cropping', restores the entire LR image before cropping the \roi{}. Both strategies are suboptimal in balancing performance versus efficiency. 
`Pre-cropping' is computationally lighter but limits restoration quality by restricting the receptive field to only the area around the ROI, resulting in suboptimal performance. `Post-cropping' provides better restoration quality by leveraging the full image context but incurs higher computational costs due to the processing of irrelevant image areas.

\begin{figure*}[t!]
    \centering
    \begin{subfigure}{0.34\textwidth}
        \centering
        \includegraphics[width=\linewidth]{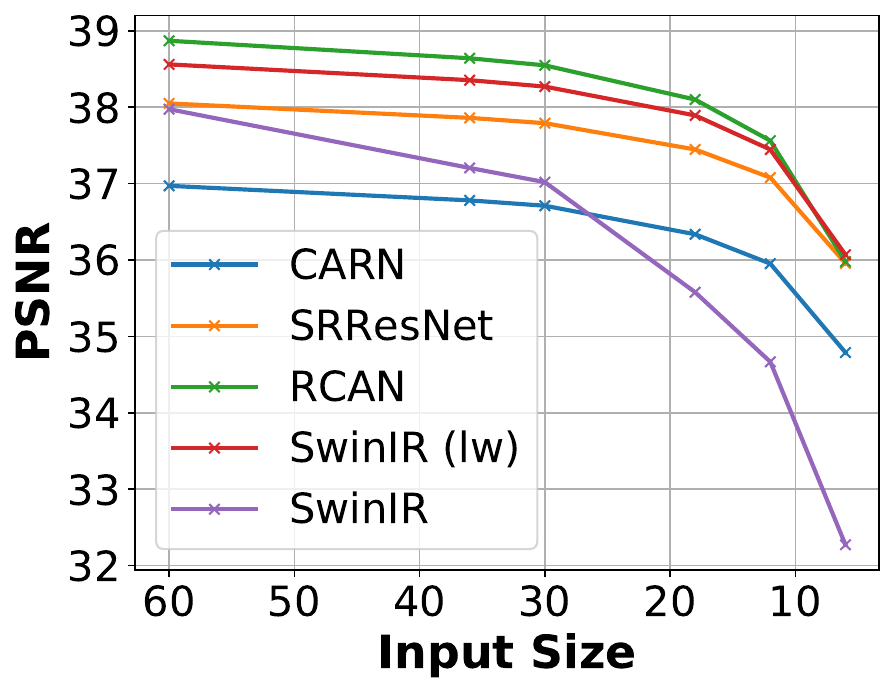}
        \subcaption{}\label{fig:inputsize_vs_psnr}
    \end{subfigure}
    \begin{subfigure}{0.55\textwidth}
        \centering
        \includegraphics[width=\linewidth]{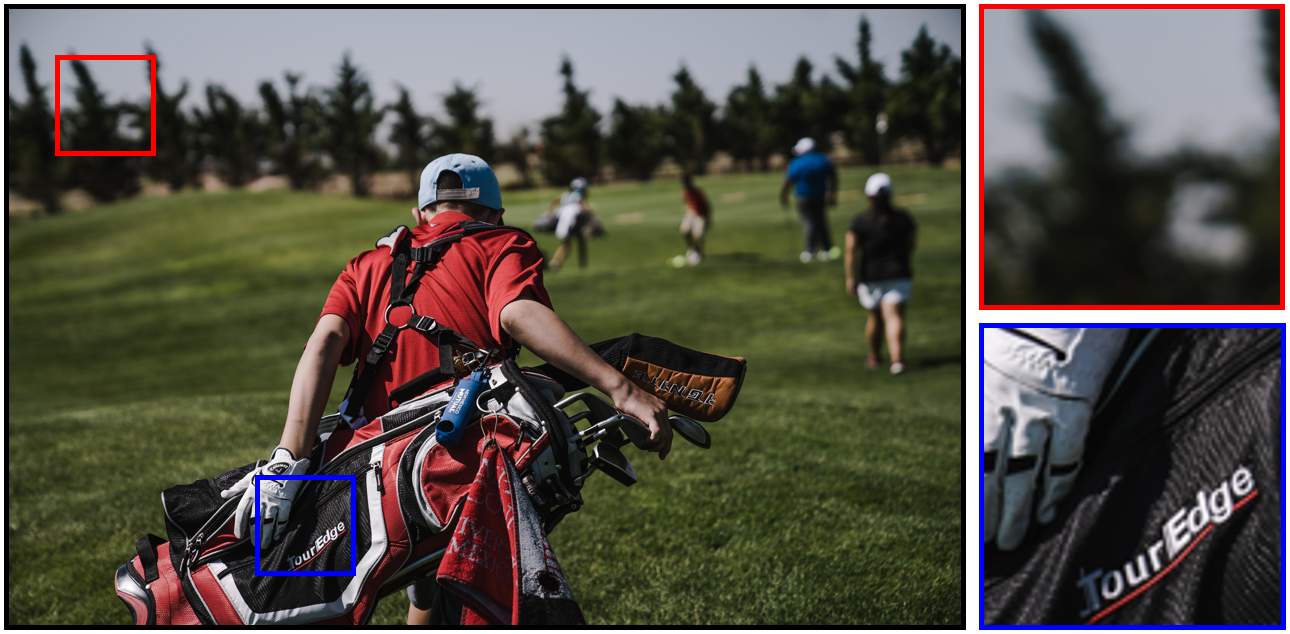}
        \subcaption{}\label{fig:patch_characteristic}
    \end{subfigure}
    \caption{
    (a) For various SR models, PSNR declines with smaller input patches. This experiment is conducted on Manga109~\cite{matsui2017sketch} dataset. (b) An example showing diverse patch characteristics including blur, detailing, and textures. 
    }\label{fig:problem_setting}
    \vspace{-0.2cm}
\end{figure*}

Without a global view from the context, it becomes challenging to discern specific details. 
`Post-cropping' performs better than `pre-cropping' precisely because it can leverage the pixels from across the image. 
By limiting the global context through reduced input size, we observe a clear relationship between input size and peak signal-to-noise ratio (PSNR), as illustrated in Figure~\ref{fig:inputsize_vs_psnr}.
As the size of the input patches decreases, every model's performance declines, with PSNRs reducing by up to 6dB.
Moreover, individual patches are diverse; some patches might appear blurry while others are clear (see Figure~\ref{fig:patch_characteristic}). 
The lack of context makes it more difficult to accurately identify and address each patch's diverse characteristics, \eg the blurriness in the trees comes from background bokeh.
The degradataion variability across different regions presents a challenge for models to perform well consistently.

To address these, we introduce \methodfull{} (\method{}) to target effective and efficient use of image context for local super-resolution. 
Our method prioritizes resources for restoring the \roi{} while expending less to extract, retrieve, and integrate information from the rest of the image.
We start with a base branch that mirrors standard SR model architectures to restore the \roi{}. 
To augment the base, we add a \globalmodule{} (\GB{}) and \localmodule{} (\LB{}) to source information from the \context{}. 
The \GB{} retrieves useful information from the entire context based on similarity.  It is not limited by the spatial distance between the \roi{} and regions in the context. 
The \LB{} is proximity-based and concentrates on integrating pixels or features in the vicinity of the \roi{}.
The \GB{} and \LB{} modules distinguish our approach from standard SR models and give our model flexibility to control the information from both global and nearby pixels for restoreing the ROI.
Our contributions are summarized as follows:
\begin{itemize}
    \item We introduce a novel task called \task{}
    to focus on restoring local regions within a LR image. We establish the training and evaluation framework for this new task. 
    \item To address \task{}, we propose \method{}, a new method which leverages context by efficiently aggregating globally and locally relevant regions around the \roi{}.
    \item 
    Our design principles for the \GB{} and \LB{} handle scenarios ROIs of arbitrary sizes and positions. 
    \item Our approach, applied to various SR backbones, outperforms both `pre-' and 'post-'cropping alternatives, demonstrates superior performance while maintaining comparable computation.
\end{itemize}

\section{Related Works} \label{sec:related_works}

\textbf{Image super-resolution} 
upsamples images and various architectures have been proposed based on convolution~\cite{lim2017enhanced,zhang2018image,chen2022simple} and transformer blocks~\cite{liang2021swinir,zamir2022restormer}.  Task variants of SR include blind SR, which investigates the impact of image degradations~\cite{zhang2018learning,xu2020unified}, video SR, which extends the concepts of SR in time~\cite{kappeler2016video,sajjadi2018frame}, perceptual-based SR, which focuses on improving the visual quality~\cite{ledig2017photo,bulat2018learn} and many more. 

There is a growing interest in accelerating SR models~\cite{zhao2020efficient,nie2021ghostsr}. Techniques like ClassSR~\cite{kong2021classsr} and APE~\cite{wang2022adaptive} increase the efficiency by using a dynamic inference budget tailored to the characteristics of different image patches. 
In this paper, our focus is on the enhancement of a specific ROI and not on the architecture itself.  
Our method \method{} is designed to be compatible with various backbones, enhancing their performance in local region restoration.\\

\noindent
\textbf{Feature matching and texture transfer} identifies and transfers relevant features or textures for various image tasks. For instance, optical flow networks ~\cite{dosovitskiy2015flownet,sun2018pwc} find local correspondences to estimate optical flow. Video object segmentation utilizes space-time memory networks to match the relevant information in the memory~\cite{oh2019video,cheng2021rethinking}. 

For SR, reference-based SR enhances an LR image using a reference HR image that shares similar content as the target image~\cite{zheng2018crossnet,shim2020robust,zhang2019image}. 
Transformers~\cite{yang2020learning} and spatial modules~\cite{lu2021masa} refine feature extraction and maintain long-range correlations. $C^2$-Matching, leveraging contrastive learning, addresses the transformation gap that often exists between LR and reference images~\cite{jiang2021robust}. Our work draws inspiration from these methodologies, viewing \task{} as a form of SR where the \context{} serves as the reference for retrieving support information. 
However, the primary distinction between our task and reference-based SR is that our reference image, namely the context, is not of high resolution and is also typically larger than the ROI. 
Moreover, we capitalize on the spatial differences and feature similarities 
between the \roi{} and the \context{} to improve the performance.

\section{Approach}\label{sec:approach}
\subsection{Formulation \& Overview}\label{sec:prelim}

Standard single-image SR, which we denote as StandardSR, restores the entire HR image $\hat{Y} \in \mathbb{R}^{3 \times (sH) \times (sW)}$ from an LR counterpart $X \in \mathbb{R}^{3 \times H \times W}$, with some fixed scalar factor $s$. The restored $\hat{Y}$ should be as close as possible the ground-truth HR image $Y \in \mathbb{R}^{3 \times (sH) \times (sW)}$. Commonly, $s$ is 2 or 4, though some methods treat $s$ as a continuous variable~\cite{wang2023deep,hu2019meta}, or apply it to factors of up to 24 or 30~\cite{cao2023ciaosr,chen2021learning}.

\task{} restores an HR version of a small region $x \in \mathbb{R}^{3 \times h \times w}$, cropped from $X$, to match the corresponding ground truth HR region $y \in \mathbb{R}^{3 \times (sh) \times (sw)}$ from $Y$. The restored HR region is denoted as $\hat{y} \in \mathbb{R}^{3\times (sh)\times (sw)}$.
It follows naturally that $h \leq H$ and $w \leq W$, though we further assume that $h \ll H$ and $w \ll W$, highlighting our focus on ROIs significantly smaller than the entire image. 
We formalize standard SR and local SR as follows:
\begin{align}
    \hat{Y} = \text{StandardSR}(X) \quad 
\text{and} \quad     \hat{y} & = \text{LocalSR}(x; X).\label{eq:local_sr}
\end{align}
The processes of Eq~\ref{eq:local_sr} are visualized in Figure~\ref{fig:local_sr} and show that StandardSR takes the LR image $X$ as input and produces the entire HR image $\hat{Y}$, whereas \task{} processes $X$ along with a specified local region $x$ (ROI) within $X$ (context), producing the corresponding HR region $\hat{y}$. In theory, the context need not span the entire LR image and its scope can vary based on application needs and image content; however, we use the entire $X$ for simplicity.

The first challenge of \task{} is the effective use of the context to restore the ROI while limiting computational complexity - balancing performance and efficiency. The `pre-' and `post-cropping' strategies
described in the introduction, \ie $\text{StandardSR}(x)$ and $\text{StandardSR}(X)$, are edge cases computationally and quality-wise for restoration. 
Our objective with $\text{LocalSR}(x; X)$ is to match or surpass the restoration quality of $\text{StandardSR}(X)$ for the specific region of $x$ while keeping the computational costs similar to $\text{StandardSR}(x)$. 
A second challenge is that the ROI can appear anywhere in the image and may vary in size. For example, the ROI could be positioned at the center, the top-left corner, or any other location. Consequently, a robust model should be able to handle ROIs of arbitrary positions and sizes within the image. This variability in ROI placement and size makes basic fusion operators like concatenation and summation, which require matching input shapes, unsuitable. Thus, a more flexible approach is needed to accommodate ROIs and contexts of differing sizes.

\begin{figure*}[t!]
    \centering
    \includegraphics[width=0.8\linewidth]{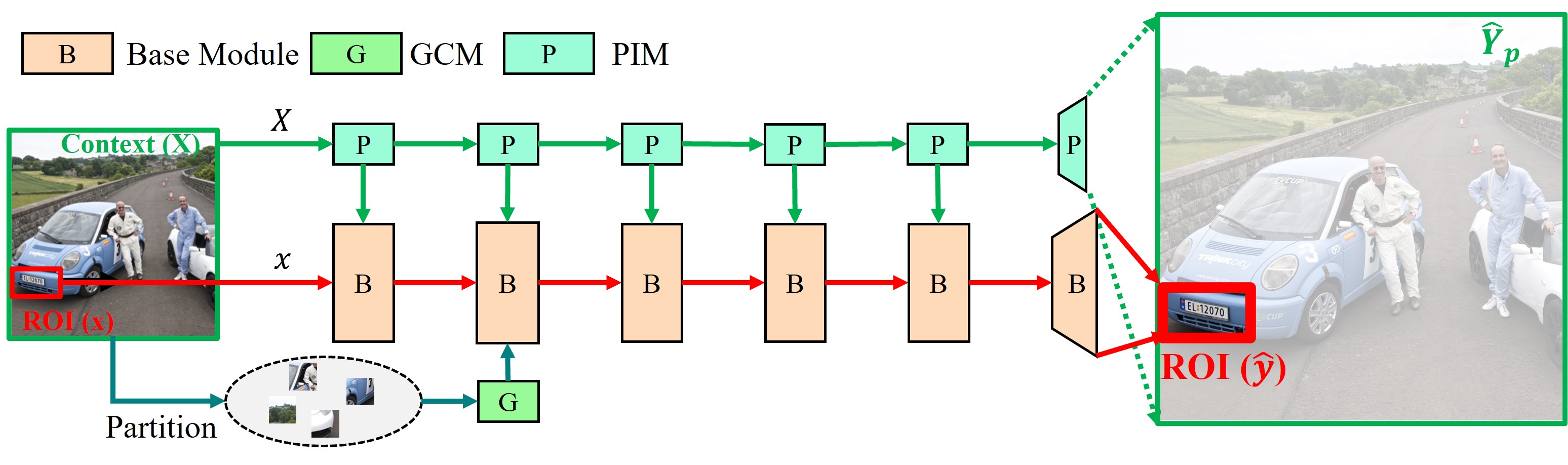}
    \caption{\textbf{The overview of \method{}.} We employ \GB{} and \LB{} to provide context features for super-resolving the ROI.  
    }
    \label{fig:overview} 
    \vspace{-0.1cm}
\end{figure*}

To address these, we develop the 
context-based local SR
(\method{}).  \method{} efficiently leverages the surrounding context but is unaffected by the ROI's relative position within the original LR image. 
It ensures adaptability and efficiency in processing diverse image scenarios.
\method{} has three branches: a base branch, a \globalbranch{} featuring \globalmodule{}s (\GB{}) (see Sec.~\ref{sec:global_branch}) and a \localbranch{} featuring \localmodule{}s (\LB{}) (see Sec.~\ref{sec:local_branch}).  Figure~\ref{fig:overview} provides an overview. 

Both the \GB{} and the \LB{} take the context, $X$, as input and generate the intermediate features for the base branch. The base branch takes the \roi{}, $x$, as input and fuses the intermediate features from the \GB{} and \LB{} at each stage to restore $\hat{y}$. 
We follow the standard image SR backbones~\cite{dong2016accelerating,ahn2018fast,zhang2018image,liang2021swinir} to design the architecture of the base branch.
The main modification is enhancing the current feature at each stage by fusing features from the \LB{} and \GB{}.
In practice, we pad the \roi{} before inputting it to the base branch for performance considerations. The output is then cropped accordingly.

\begin{figure}[t!]
    \centering
    \begin{subfigure}{0.9\linewidth}
        \includegraphics[width=\linewidth]{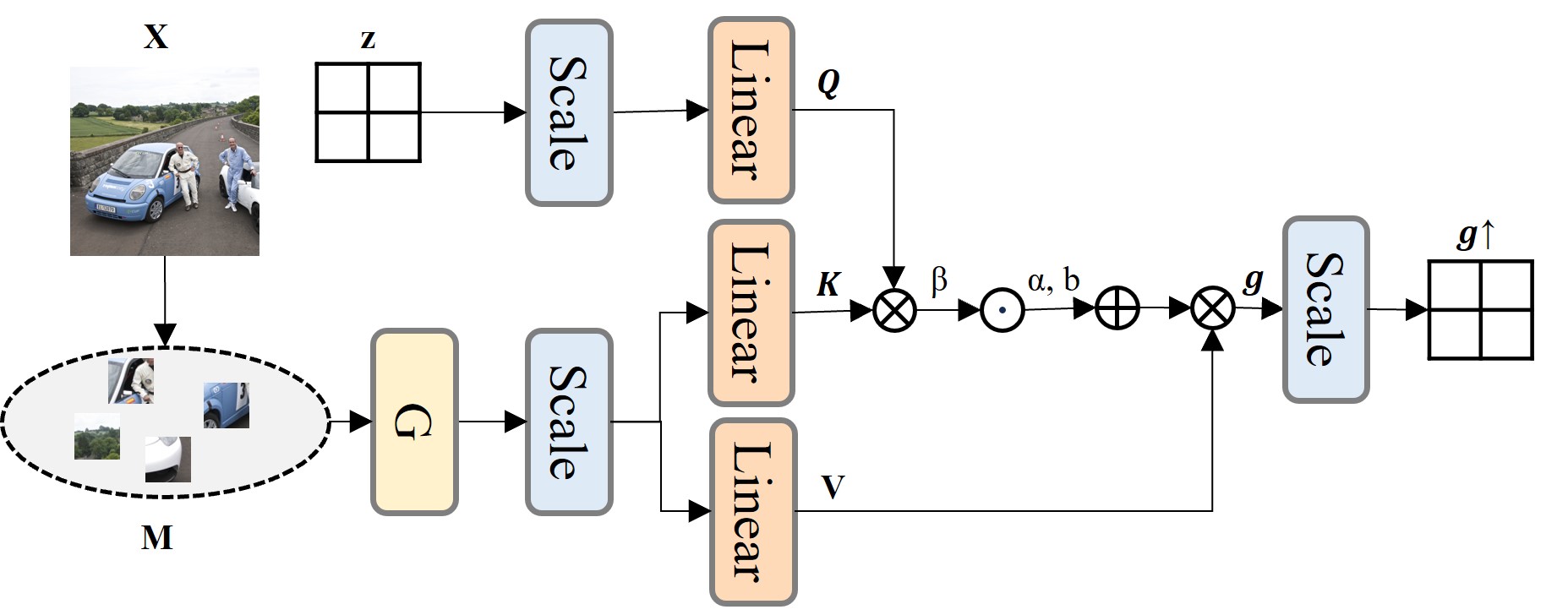}
        \subcaption{\capfirstglobalmodule{} (\GB{}).}
        \label{fig:global_branch}
    \end{subfigure}
    \begin{subfigure}{0.9\linewidth}
        \includegraphics[width=\linewidth]{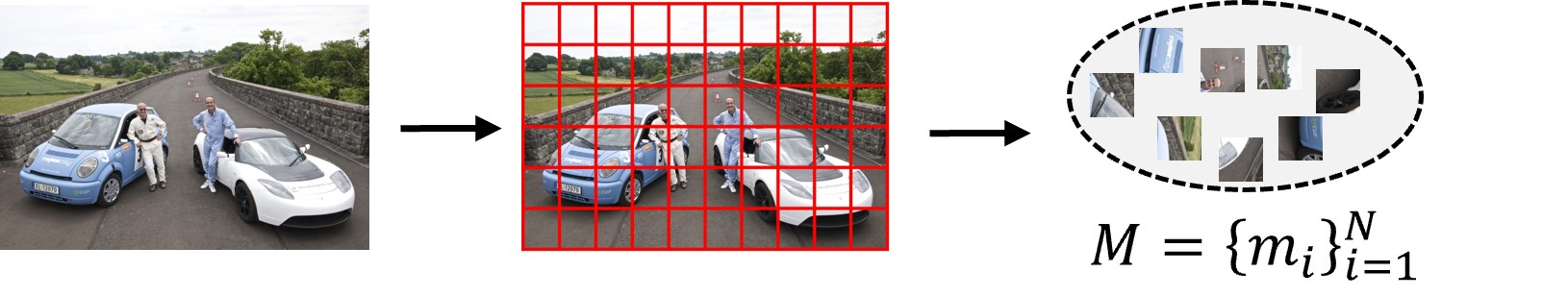}
        \subcaption{Partition the context into a set of features.}
        \label{fig:global_branch_partition}
    \end{subfigure}
    \begin{subfigure}{0.9\linewidth}
        \includegraphics[width=\linewidth]{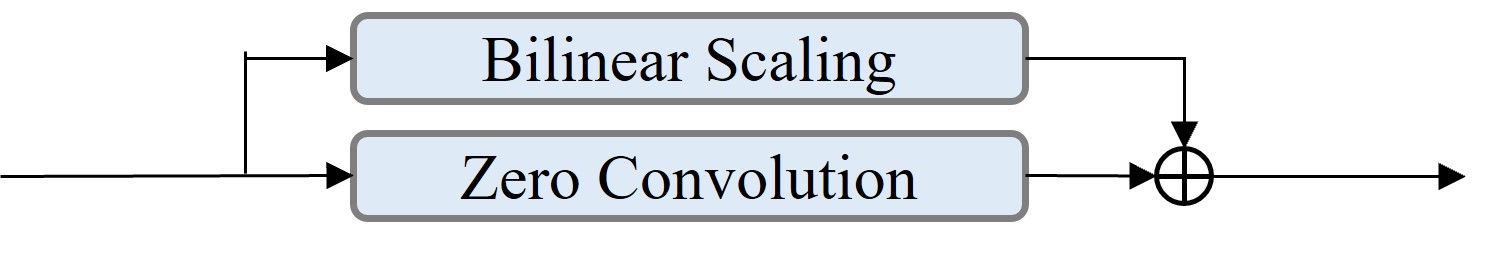}
        \subcaption{The scale module in \GB{}.}
        \label{fig:scale}
    \end{subfigure}
    \caption{\textbf{\capfirstglobalmodule{} (\GB{}).} The architecture of \GB{} is shown in (a), with its partition strategy detailed in (b) and scale module in (c).}
    \vspace{-0.5cm}
\end{figure}

\subsection{\capfirstglobalmodule~(\GB{})}\label{sec:global_branch}

The \globalmodule{} (\GB{}) allows the \roi{} to retrieve useful information from the entire context, unrestricted by the distance. 
We seek a method to selectively aggregate important pixels within the context, regardless of their positions or sizes. This is handled more effectively by similarity-based attention mechanisms than standard convolutions.
Inspired by cross-attention~\cite{vaswani2017attention}, we partition the context image into non-overlapping patches and use the ROI's query features to retrieve matching context features. This approach outputs features of the same shape as the ROI, allowing fusion through basic operations like summation and concatenation, which requires the two inputs to be of the same shape.
Figure~\ref{fig:global_branch} shows the overview of \GB{}.

Useful features from the context are collected by the partition strategy illustrated in Figure~\ref{fig:global_branch_partition}.
Specifically, we divide the context $X$ into patches of size $r\times r$.
Since the computational cost of querying patches increases with the number of patches, we subsample $N$ representative patches to boost efficiency.
Then, a feature extractor module $\mathcal{G}$ transforms each sampled RGB patch into a feature, creating a collection of $N$ features, denoted as $M:=\{m_i\}_{i=1}^N$. 
Each feature $m_i$ corresponds to a specific local window within $X$. These features are independent of each other in the subsequent operations. 
In our implementation, we apply a non-overlapping partitioning strategy. 
The subset is chosen by uniformly sampling along the spatial dimensions, such that $N$ is less than $(H/r)\! \times\! (W/r)$.  Such a subsampling strategy is simple yet effective.

The resulting features $M$ are then transferred to the base branch 
to enhance the super-resolution of the ROI.
Let $z$ represent an intermediate feature from the base branch.
The feature $z$ and each element $m_i$ in $M$ are downsampled into $(z \downarrow)$ and $(m_i\downarrow)$. Downsampling curtails computational and memory demands by decreasing tokens and reducing noise.
Then, one projection layer with multiple heads maps $(z\downarrow)$ into \textit{query} and $(m_i\downarrow)$ into a \textit{key} and \textit{value} (denoted as $Q$, $K$ and $V$ respectively):
\begin{align}
    \!\! Q=(z\downarrow)W_Q, \;K=(M\downarrow)W_K,\; V=(M\downarrow)W_V,\label{eq:downsample}
\end{align}
\noindent where $W_Q,W_K,W_V\in \mathbb{R}^{c\times c}$ and $(M\!\downarrow)=\{(m_i\!\downarrow)\}_{i=1}^N$. The aggregation is:
\begin{align}
    g = \text{softmax}(\alpha + \beta \cdot {QK^T} / \sqrt{c} + b)\cdot V,\label{eq:qkv},
\end{align}
where $\alpha$ and $\beta$ are learnable scalar parameters to scale and shift the similarity, $g$ is the aggregated result, and the $b$ is a bias constant to emphasize spatial proximity. The value of $b$ is computed as the pixel distance of the query and key in the LR image.  
$g$ is then scaled up by an upsampler. \\

\noindent
\textbf{Zero convolution with bilinear scaling.} 
Our downsampler and upsampler consist of convolutional layers with a bilinearly scaled residual. The convolutions feature learnable parameters to improve expressive power. However, as we use a residual setup, randomly initializing such convolutions may lead to the loss or blurring of crucial details during the downsampling and upsampling stages. 
We therefore propose to initialize the convolutions in the downsampler and upsampler with zeros, inspired by ControlNet~\cite{zhang2023adding}. 
This initialization strategy ensures that, after adding the residual, the output from these newly initialized modules resembles a bilinearly interpolated image.
The architecture of this approach is depicted in Figure~\ref{fig:scale}. 
The formulation with respect to the input $z$ and $g$ is as follows; it is similarly applied to  $M$ for the downsampling. 
\begin{align}
    (z\downarrow) &  = \mathcal{S}_{down}(z;\theta_{down}) + (z\downarrow_{BI});  \\
    (g\uparrow) & = \mathcal{S}_{up}(g;\theta_{up}) + (g\uparrow_{BI}),
\end{align}
where $\mathcal{S}_{down}$ and $\mathcal{S}_{up}$ represent the scaling convolution module in the downsampler and upsampler with corresponding trainable parameters $\theta_{down}$ and $\theta_{up}$, $\downarrow_{BI}$ and $\uparrow_{BI}$ denote the bilinear scaling. $\theta_{down}$ and $\theta_{up}$ are initialized as zeros, allowing the network to initially rely on bilinear scaling for stable output in early training, while gradually adjusting the zero-initialized branch for optimized performance.
$(g\uparrow)$ is the final output of the \GB{}.

\begin{figure}[b!]
    \centering
    \includegraphics[width=\columnwidth]{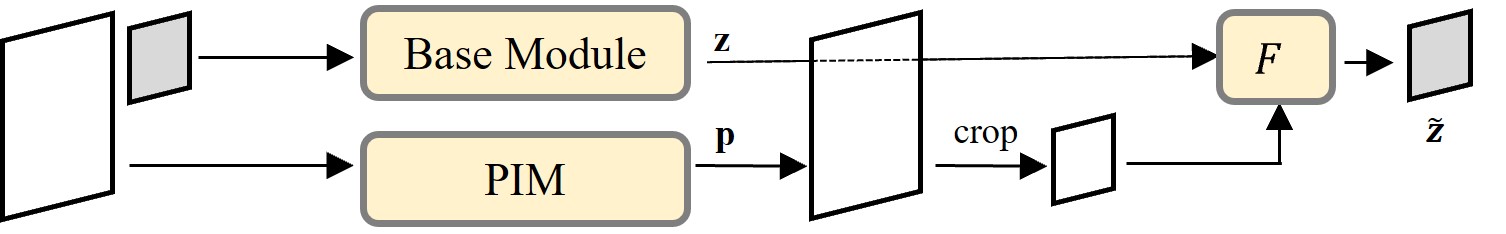}
    \caption{\textbf{\capfirstlocalmodule{} (\LB{})}. The grey and white squares represent the features in the base branch and the \localbranch{}, respectively.}
    \label{fig:local_branch}
\end{figure}

\subsection{\capfirstlocalmodule{} (\LB{})}\label{sec:local_branch}
Since pixels closer to the ROI are more likely to contribute to its restoration, this should be reflected in pipeline. Rather than encoding distance information directly into \GB{}, which could shift the model's focus from feature similarity to spatial distance, we introduce an independent branch that mainly aggregates features surrounding the ROI.
This branch consists of multiple \localmodule{}s (\LB{}) arranged in series, as illustrated in Figure~\ref{fig:overview}, where each instance of `P' represents an \LB{}, collectively forming the branch.
The \LB{} shares a similar architecture as the base module but has fewer channels in the convolutions or transformer blocks to save computational costs.

This branch can be viewed as a slim version of the base branch, yet it processes the context instead of the ROI. 
As such, it produces its own output, denoted as $\hat{Y}_{P}$, which represents the HR version of the context image. This output is of the same spatial size as the context, \ie $H\!\times\!W$, and is used to supervise the model training, as shown in Eq~\ref{eq:loss}.
When fusing features from \LB{}, we first crop the corresponding ROI region from the features in \LB{} to align with the ROI's shape. 
This cropped context feature is then concatenated with the ROI features and passed to a fusion operator, $\mathcal{F}$.
Since the cropped regions are derived from a local receptive field within \LB{}'s features, their aggregation effectively collects pixels near the ROI.

Specifically, the feature from the \LB{} is integrated to the base branch with fusion operation $\mathcal{F}$ to derive the output $\tilde{z}$: 

\begin{align}
    \tilde{z} = \mathcal{F}(z, \text{crop}(p)).\label{eq:local_branch}
\end{align}
Above, $z \in \mathbb{R}^{c\times h\times w}$ is the feature from the base branch, while $p \in \mathbb{R}^{c'\!\times\! H\!\times\! W}$ is the feature from the \LB{} corresponding spatially to the \roi{}. Note that $c'\! <\! c$, as the \LB{} has fewer channels. 
The fusion operation $\mathcal{F}$ 
is implemented as a sum across the first $c'$ channels out of consideration for simplicity. A simple illustration is in Figure~\ref{fig:local_branch}.

\subsection{Complexity analysis}\label{sec:approach_analysis}
Both \GB{} and \LB{} work well regardless of the relative positioning and size of the ROI with respective to the entire image.
In the \GB{}, the context is divided into independent patches, queried equally without shape restrictions. The \LB{} processes the entire context, allowing for the simple cropping of the corresponding region in the feature. 

\begin{table*}[t!]
  \centering
  \setlength{\tabcolsep}{5pt}
  \resizebox{\textwidth}{!}{
  \begin{tabular}{c|c|c|ccc|ccc|ccc|ccc}
    \toprule
    & \multirow{2}{*}{Method} & \multirow{2}{*}{Variant}  & \multicolumn{3}{c|}{BSD100} & \multicolumn{3}{c|}{Urban100} & \multicolumn{3}{c|}{Manga109} & \multicolumn{3}{c}{Test2K}   \\
    & & &   PSNR & SSIM & GFLOPs &  PSNR & SSIM & GFLOPs & PSNR & SSIM & GFLOPs & PSNR & SSIM & GFLOPs  \\
    \midrule 
  \parbox[t]{2mm}{\multirow{9}{*}{\rotatebox[origin=c]{90}{CNN-based}}} & \multirow{3}{*}{CARN~\cite{ahn2018fast}}& \baselineOne{} & 27.27 & 0.8839 & 0.65 & 25.09 & 0.8852 & 0.65 & 28.82 & 0.9574 & 0.65 & 27.23 & 0.8893 & 0.65 \\
 & & \baselineTwo{} & 27.38 & 0.8854 & 35.37 & 25.29 & 0.8879 & 209.59 & 29.27 & 0.9597 & 267.23 & 27.34 & 0.8910 & 489.93 \\
 & & Ours & \color{blue}\textbf{27.39} & \color{blue}\textbf{0.8854} & 1.63 & \color{blue}\textbf{25.32} & \color{blue}\textbf{0.8883} & 5.65 & \color{blue}\textbf{29.24} & \color{blue}\textbf{0.9595} & 6.97 & \color{blue}\textbf{27.34} & \color{blue}\textbf{0.8909} & 12.11 \\
\cmidrule{2-15}
 & \multirow{3}{*}{SRResNet~\cite{ledig2017photo}}& \baselineOne{} & 27.51 & 0.8880 & 2.93 & 25.77 & 0.8983 & 2.93 & 29.74 & 0.9640 & 2.93 & 27.44 & 0.8939 & 2.93 \\
& & \baselineTwo{} & 27.63 & 0.8895 & 158.19 & 25.99 & 0.9013 & 937.44 & 30.19 & 0.9658 & 1195.23 & 27.56 & 0.8957 & 2191.26 \\
 & & Ours & \color{blue}\textbf{27.62} & \color{blue}\textbf{0.8894} & 5.08 & \color{blue}\textbf{26.03} & \color{blue}\textbf{0.9018} & 8.94 & \color{blue}\textbf{30.20} & \color{blue}\textbf{0.9656} & 10.22 & \color{blue}\textbf{27.57} & \color{blue}\textbf{0.8957} & 15.15 \\
\cmidrule{2-15}
 & \multirow{3}{*}{RCAN~\cite{zhang2018image}}& \baselineOne{} & 27.67 & 0.8909 & 18.38 & 26.35 & 0.9087 & 18.38 & 30.47 & 0.9682 & 18.38 & 27.60 & 0.8971 & 18.38 \\
 & & \baselineTwo{} & 27.82 & 0.8926 & 992.73 & 26.67 & 0.9126 & 5882.85 & 31.07 & 0.9707 & 7500.64 & 27.76 & 0.8991 & 13751.17 \\
 & & Ours & 27.79 & 0.8924 & 19.41 & 26.66 & 0.9124 & 21.50 & 31.05 & 0.9706 & 22.19 & 27.75 & 0.8990 & 24.86 \\
\midrule
 \parbox[t]{2mm}{\multirow{6}{*}{\rotatebox[origin=c]{90}{Transformer-based}}}  & \multirow{3}{*}{SwinIR (lw)~\cite{liang2021swinir}} & \baselineOne{} & 27.63 & 0.8901 & 2.11 & 26.09 & 0.9045 & 2.11 & 30.26 & 0.9673 & 2.11 & 27.56 & 0.8962 & 2.11 \\
 & & \baselineTwo{} & 27.76 & 0.8917 & 113.70 & 26.37 & 0.9082 & 673.77 & 30.79 & 0.9693 & 859.06 & 27.71 & 0.8982 & 1574.94 \\
 & & Ours & \color{blue}\textbf{27.76} & \color{blue}\textbf{0.8918} & 3.18 & \color{blue}\textbf{26.38} & \color{blue}\textbf{0.9084} & 6.74 & \color{blue}\textbf{30.80} & \color{blue}\textbf{0.9692} & 7.92 & \color{blue}\textbf{27.71} & \color{blue}\textbf{0.8982} & 12.47 \\
\cmidrule{2-15}
 & \multirow{3}{*}{SwinIR~\cite{liang2021swinir}}& \baselineOne{} & 27.75 & 0.8923 & 15.08 & 26.53 & 0.9118 & 15.08 & 30.76 & 0.9706 & 15.08 & 27.68 & 0.8987 & 15.08 \\
&  & \baselineTwo{} & 27.90 & 0.8942 & 814.40 & 26.90 & 0.9165 & 4826.06 & 31.36 & 0.9728 & 6153.23 & 27.86 & 0.9011 & 11280.91 \\
 & & Ours & \color{blue}\textbf{27.90} & \color{blue}\textbf{0.8942} & 23.23 & \color{blue}\textbf{26.91} & \color{blue}\textbf{0.9165} & 31.02 & \color{blue}\textbf{31.36} & \color{blue}\textbf{0.9728} & 33.60 & 27.85 & 0.9010 & 43.56 \\
    \bottomrule
  \end{tabular}
  }
\caption{\textbf{Quantitative comparison on different backbones with ROI size as $24\!\times\! 24$.} Our method outperforms the Pre-cropping in PSNR with only a slight increase in FLOPs and achieves comparable or superior performance to the Post-cropping method while significantly reducing FLOPs. We apply our approach to CNN-based networks (CARN~\cite{ahn2018fast}, SRResNet~\cite{ledig2017photo},  RCAN~\cite{zhang2018image}) and to transformer-based networks (SwinIR~\cite{liang2021swinir}).
}
  \label{table:comparison_baseline_rs24_x4}
\end{table*}

\begin{figure*}
    \centering
    \begin{subfigure}{0.33\textwidth}
        \centering
       \includegraphics[width=\linewidth]{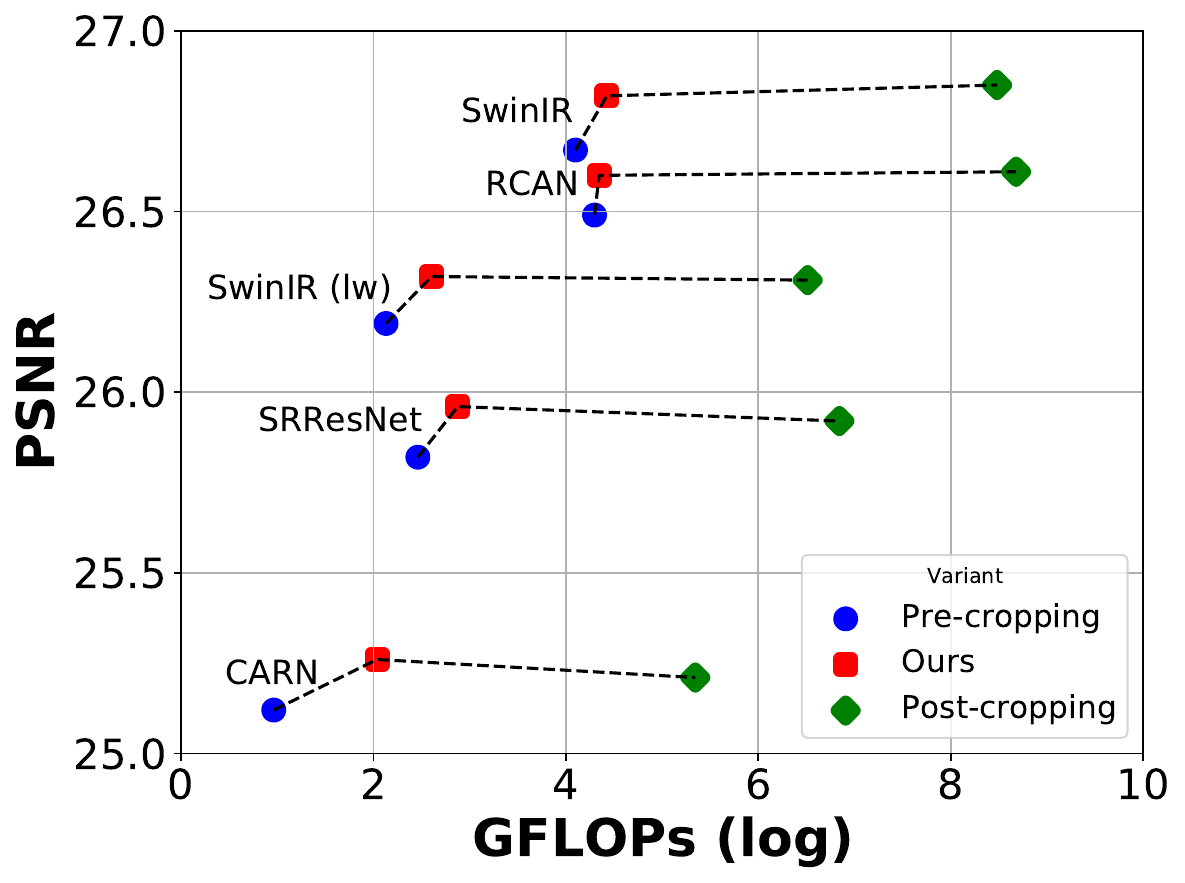}
        \subcaption{Urban100}\label{}
    \end{subfigure}
    \begin{subfigure}{0.33\textwidth}
        \centering
       \includegraphics[width=\linewidth]{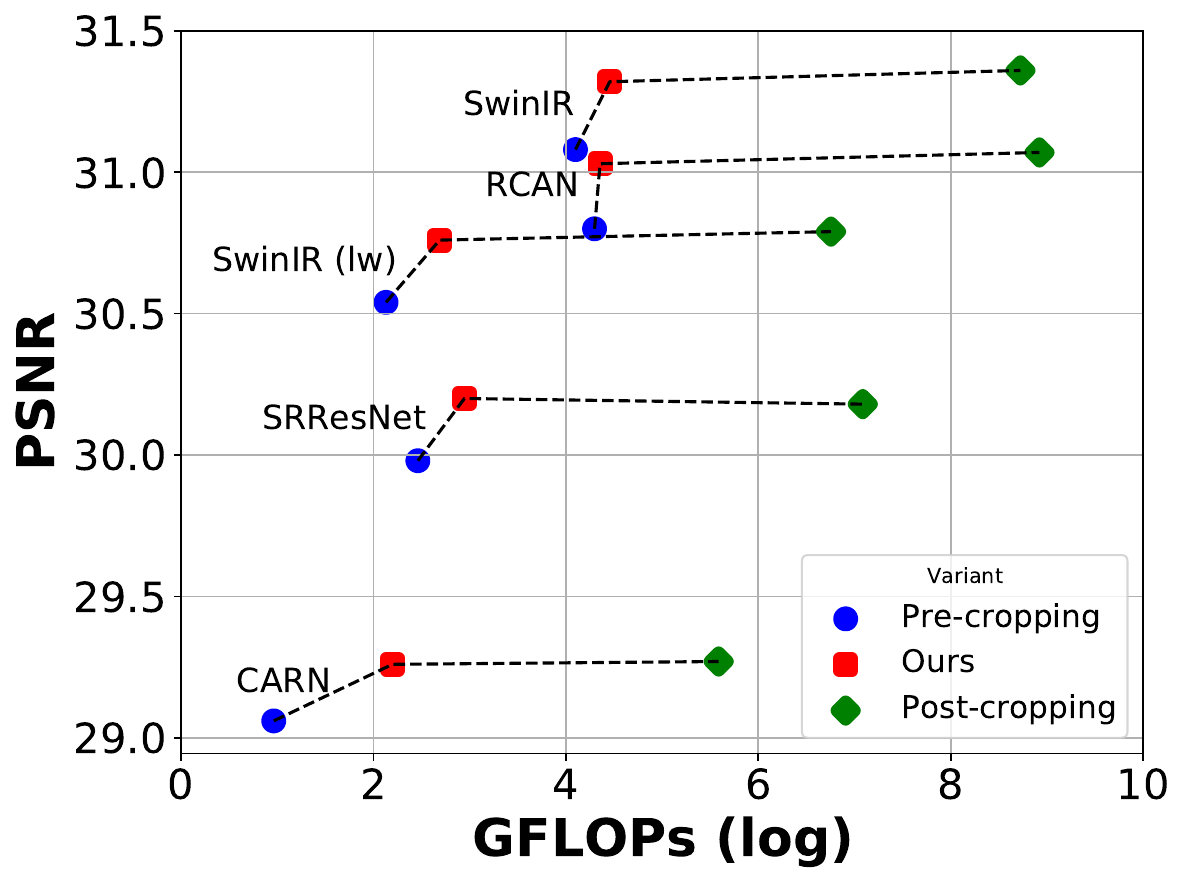}
        \subcaption{Manga109}\label{}
    \end{subfigure}
    \begin{subfigure}{0.33\textwidth}
        \centering
       \includegraphics[width=\linewidth]{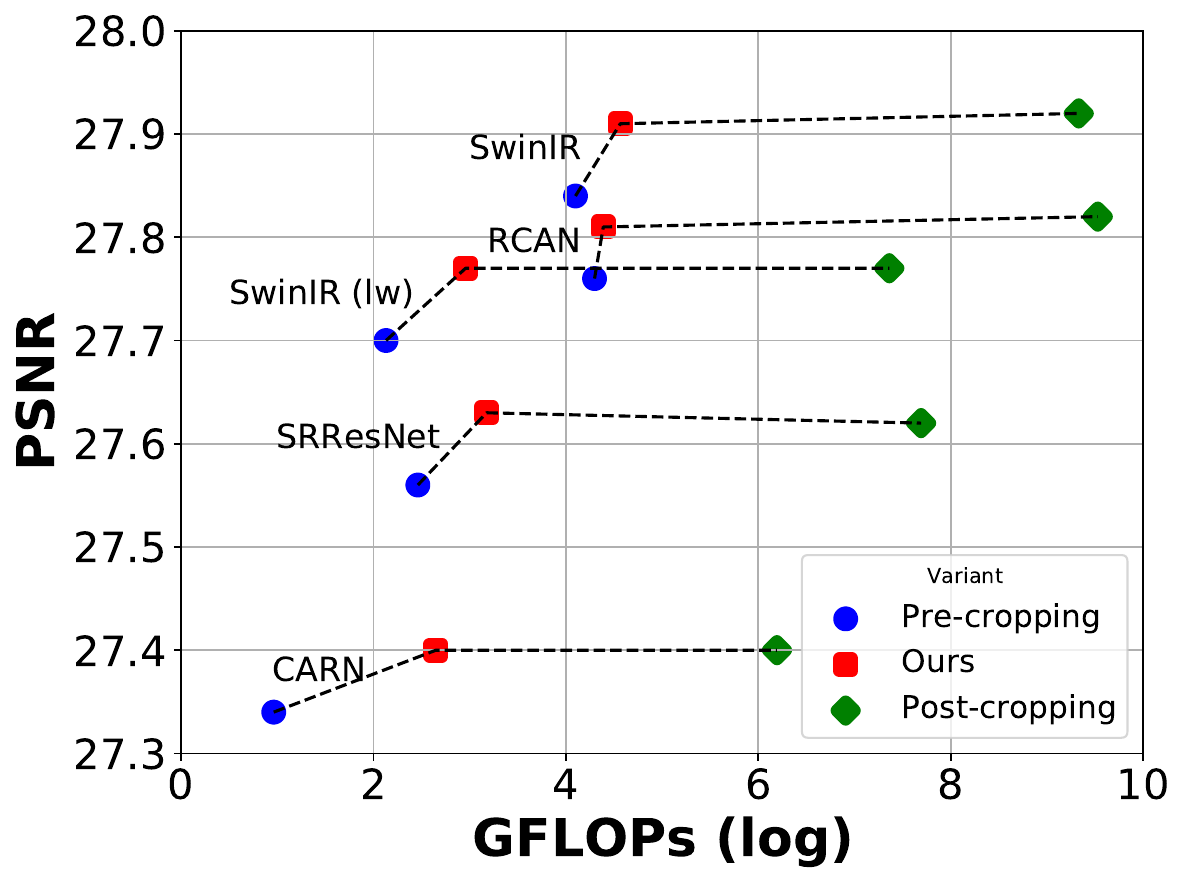}
        \subcaption{Test2K}\label{}
    \end{subfigure}
    \caption{\textbf{Quantitative comparison on different backbones with ROI size as $48\!\times\!48$.} 
    Dashed lines connect experiments using the same backbone, with labels on the left. {\color{red} Our method} outperforms `{\color{blue}pre-cropping}' in PSNR with a slight FLOPs increase and achieves comparable or superior performance to `{\color{ForestGreen}post-cropping}' while significantly reducing FLOPs.
    }\label{fig:sota_48x48}
\end{figure*}
\begin{figure*}
    \centering
    \begin{subfigure}{0.24\textwidth}
        \centering
        \includegraphics[width=\linewidth]{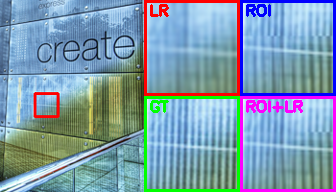}
        \subcaption{}\label{fig:vc_1}
    \end{subfigure}
    \begin{subfigure}{0.24\textwidth}
        \centering
        \includegraphics[width=\linewidth]{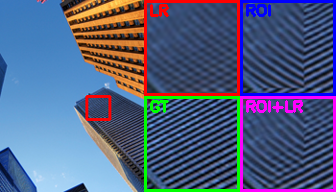}
        \subcaption{}\label{fig:vc_2}
    \end{subfigure}
    \begin{subfigure}{0.24\textwidth}
        \centering
        \includegraphics[width=\linewidth]{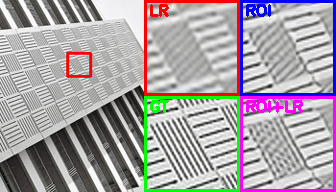}
        \subcaption{}\label{fig:vc_3}
    \end{subfigure}
    \begin{subfigure}{0.24\textwidth}
        \centering
        \includegraphics[width=\linewidth]{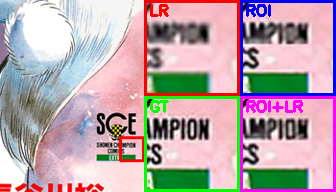}
        \subcaption{}\label{fig:vc_4}
    \end{subfigure}

    \begin{subfigure}{0.24\textwidth}
        \centering
        \includegraphics[width=\linewidth]{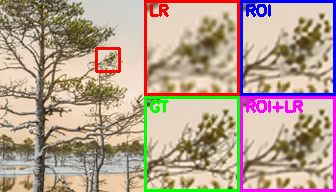}
        \subcaption{}\label{fig:vc_5}
    \end{subfigure}
    \begin{subfigure}{0.24\textwidth}
        \centering
        \includegraphics[width=\linewidth]{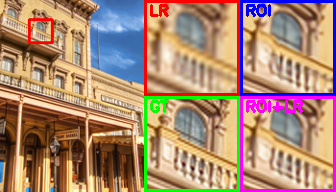}
        \subcaption{}\label{fig:vc_6}
    \end{subfigure}
    \begin{subfigure}{0.24\textwidth}
        \centering
        \includegraphics[width=\linewidth]{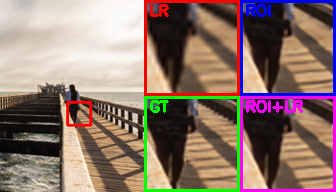}
        \subcaption{}\label{fig:vc_7}
    \end{subfigure}
    \begin{subfigure}{0.24\textwidth}
        \centering
        \includegraphics[width=\linewidth]{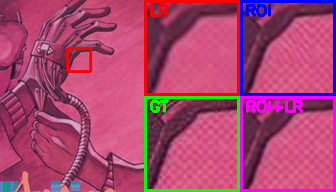}
        \subcaption{}\label{fig:vc_8}
    \end{subfigure}
    \vspace{-0.2cm}
    \caption{\textbf{Visual comparison with the baseline competitors.} Four sub images represent the LR input, ``ROI'', ``ROI+LR'' and ground-truth HR image, respectively. ``ROI'' is the baseline, and ``ROI+LR'' is our proposal which uses LR as the context.}\label{fig:visual_comparison}
    \vspace{-0.2cm}
\end{figure*}

Image SR models generate features for each pixel in the input, with the complexity of processing the entire context image $X \in \mathbb{R}^{H \times W}$ expressed as $O(H \times W)$, which scales quadratically with image size. This computational burden grows further when the context size increases.
In our model, the complexity for restoring a single region is $O(hw) + O_G(HW) + O_P(HW)$, where $O(hw)$, $O_G(HW)$, and $O_P(HW)$ represent the complexities of the base branch, \globalbranch{}, and \localbranch{}, respectively.
The \LB{} and \GB{} process the \context{} independently from the base branch, so there are additional compute gains, from both parallel processing and from re-use of pre-computed feature maps for restoring multiple regions of the same image.  
To restoring $n$ patches in the same image, the complexity is $nO(hw)\!+\!O_G(HW)\!+\!O_P(HW)$, 
showcasing the efficiency of our approach in handling multiple patches.

\subsection{Training, Inference, \& Loss}\label{sec:training_and_evaluation}

\textbf{Training.} In principle, standard SR methods should be trained with entire LR/HR pairs. However, most methods~\cite{ahn2018fast,ledig2017photo,zhang2018image,liang2021swinir,zhang2019deep}, due to hardware memory limitations, are trained by sampling overlapping patches from the LR/HR pair. 
Similarly, hardware memory prevents the use of the entire image $X$ as the context for \task{}.  We thus use sampled LR/HR patches for training, where the patch is treated as the context, and the center region as the \roi{} $x$.

\noindent \textbf{Inference.} In standard SR, inference and evaluation is applied to the entire image $X$. For \task{}, we divide the image into non-overlapping patches, treat each as the \roi{} $x$, and the entire LR image as the context.
All experiments are conducted patch-wise, with the ROI located flexibly across the image; it can be centered within the image or positioned near the edges or corners.

\noindent \textbf{Loss.} We use the following training loss:
\begin{align}
    \mathcal{L} = \|\hat{y} - y\|_1 + \lambda \|\hat{Y}_{P} - Y\|_1,\label{eq:loss}
\end{align}
where $y$ and $Y$ denote the ground-truth high-resolution \roi{} and \context{}, and $\hat{y}$
and $\hat{Y}_P$ are the restored \roi{} and restored context from \LB{}.  $\lambda$ is a weighting hyperparameter. The first loss term focusing on the \roi{} supervises the entire model.  The second loss term is applied to the entire image $Y$.  While recovering $Y$ is not our objective, this term is necessary to train the \localbranch{}, since the gradients from the first loss term are  
concentrated only in the cropped region processed by the \LB{} and other regions have no direct supervision. 
In our experiments, we set $\lambda\!=\! 0.5$ initially, then gradually reduce it to zero to allow the model to focus more on \roi{} restoration.

\section{Experiments}\label{sec:exp}

\subsection{Setting}
\noindent \textbf{Architecture.} 
We apply our approach to convolution-based networks, including CARN~\cite{ahn2018fast}, SRResNet~\cite{ledig2017photo},  RCAN~\cite{zhang2018image}, and to transformer-based networks such as SwinIR~\cite{liang2021swinir}. 
SRResNet, CARN and lightweight SwinIR are considered 
efficient architectures while RCAN and SwinIR are 
deep SR models. We set local window size $r\!=\!6$ 
and the channels of the \LB{} as $c'\!=\!c/10$, empirically. 
The feature extraction module $\mathcal{G}$ in the \GB{} is based on the first few stages of each selected architecture. $\mathcal{S}_{down}$ is implemented as a single-layer strided convolution. $\mathcal{S}_{up}$ employs a single-layer transposed convolution. 
To enhance efficiency, integration from the \GB{} to the base module occurs only once per backbone, thereby reducing the number of required matching operations.

\noindent \textbf{Training details.}
The training dataset is the DIV2K~\cite{agustsson2017ntire}.The model is initialized with pre-trained weights and optimized for 200k iterations. The learning rate is set as $1\!\times\!e^{-4}$ with cosine learning rate decay strategy.

\begin{table}[t!]
  \centering
  \begin{tabular}{cccc}
    \toprule
    Period& Dataset & \roi{} & \Context{} \\
    \midrule
    Training & DIV2K& $48\!\times\!48$ &  $54\!\times\!54$ \\ 
    \midrule
    \multirow{4}{*}{Evaluation} & BSD100 & \multirow{2}{*}{$24\!\times\!24$} & $144\!\times\!216$\\ 
    & Urban100 & \multirow{2}{*}{\&} & $384\!\times\!480$\\
    & Manga109 &\multirow{2}{*}{$48\!\times\!48$} & $576\!\times\!408$\\ 
    & Test2K &  & $528\!\times\!816$ \\
    \bottomrule
  \end{tabular}
  \caption{\textbf{Average ROI and context sizes in our implementation.}} \label{table:roi_context_setup}
  \vspace{-0.2cm}
\end{table}

\noindent 
\textbf{Evaluation details.} The evaluation datasets include 
BSD100~\cite{martin2001database}, Manga109~\cite{matsui2017sketch}, Urban100~\cite{huang2015single}, Test2K~\cite{agustsson2017ntire,kong2021classsr}. 
To comply with the input size requirements of all models, images are appropriately cropped. For instance, SwinIR requires image sizes to be multiples of its window size; therefore, images are cropped to satisfy this criterion. 
We report PSNR and SSIM~\cite{wang2004image}. 
Unlike previous works~\cite{liang2021swinir,lim2017enhanced}, we do not crop the borders of each patch when calculating PSNR and SSIM as this goes against the spirit of our work in recovering a given ROI. Moreover, SSIM is calculated between restored patches instead of whole images, resulting in higher SSIM values in our reported tables. The complexity is assessed with FLOPs. 
We evaluate our model by setting the ROI size to $24 \times 24$ and $48\!\times\! 48$, with the context being the original LR image. 
We present results on $\times4$ SR.
We list the average sizes of the ROIs and contexts used in our implementation in Table~\ref{table:roi_context_setup}.

\subsection{Comparison with the baselines}
Table~\ref{table:comparison_baseline_rs24_x4} and Figure~\ref{fig:sota_48x48} present comparisons between our method and the `pre-cropping' and `post-cropping' baselines, with ROI sizes set to $24 \times 24$ and $48 \times 48$, respectively, using the original LR input as context. Padding is considered in this comparison.

For $24 \times 24$ \roi{}s (Table~\ref{table:comparison_baseline_rs24_x4}), our approach achieves gains of $0.2-0.6$dB compared to the `pre-cropping' baseline.  The PSNRs are actually comparable or exceeding the `post-cropping' baseline, all while using only a mere $0.1-1.9\%$ of the FLOPS of the latter.
This efficiency is achieved because `post-cropping' expends resources to restore unnecessary regions while our method prioritizes the restoration of the local region.

For $48\!\times\!48$ ROIs, we use figures, rather than tables, to present the results, as shown in Figure~\ref{fig:sota_48x48}. These figures offers a more intuitive comparison. 
Dashed lines connect experiments using the same backbone, with annotations on the left. {\color{red} Our proposal} outperforms the `{\color{blue}pre-cropping}' baseline in PSNR with a slight increase of FLOPs and achieves comparable or superior performance to the `{\color{ForestGreen}post-cropping}' baseline, while significantly reducing FLOPs.
A larger ROI size narrows
the PSNR performance gap between our approach and the `pre-cropping' baseline. 
The maximum performance gain is 0.24dB on Manga109. 

Combining the results from Tables~\ref{table:comparison_baseline_rs24_x4} and Figure~\ref{fig:sota_48x48}, we conclude that our approach is more effective when the ROI is significantly smaller than the original image.
Manga109 shows the most substantial impact from a lack of contextual information compared to other datasets. 
This is likely attributed to Manga109's high prevalence of 
details
, such as straight lines, that are inherently more sensitive to the surrounding context.

Figure~\ref{fig:visual_comparison} shows a visual comparison. 
Figures \ref{fig:vc_1}--\ref{fig:vc_3} 
demonstrate that the baseline ``ROI'' variant (top right subimage) tends to misalign straight lines more frequently compared to our method (bottom right subimage). Meanwhile, Figures \ref{fig:vc_4}--\ref{fig:vc_5} illustrate that the baseline produces blurrier patterns. Our proposal, leveraging the context information, significantly improves these patterns despite the input ROI being a small patch size.
Figures \ref{fig:vc_6}--\ref{fig:vc_8} highlight the presence of blurry artifacts around lines, especially near different objects or patterns. These visual comparisons demonstrate the difficulty in achieving accurate restoration with a patch-limited approach. Our method mitigates this by considering other pixels in the image.

\begin{table}[t!]
  \centering
    \begin{tabular}{cccc}
    \toprule
    \LB{} & \GB{} & PSNR & SSIM \\
    \midrule
    \xmark & \xmark & 37.64 &  0.9927 \\ 
    \cmark & \xmark & 37.93 & 0.9928\\ 
    \xmark & \cmark &  37.83 & 0.9929\\ 
    \cmark & \cmark & \textbf{37.99} & \textbf{0.9929}\\ 
    \bottomrule
  \end{tabular}
  \caption{\textbf{Ablation study on proposed modules.} Combining \GB{} and \LB{} yields the best result.} \label{table:abl_modules}
  \vspace{-0.5cm}
\end{table}

\begin{figure}
    \centering
    \begin{subfigure}{0.47\linewidth}
        \centering
        \includegraphics[width=\linewidth]{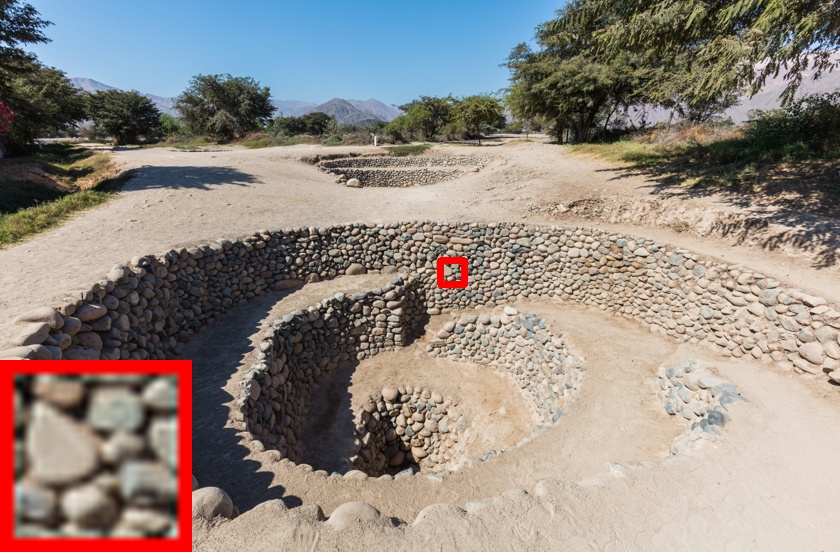}
        \subcaption{Input and context}\label{fig:vis_input}
    \end{subfigure}
    \begin{subfigure}{0.47\linewidth}
        \centering
        \includegraphics[width=\linewidth]{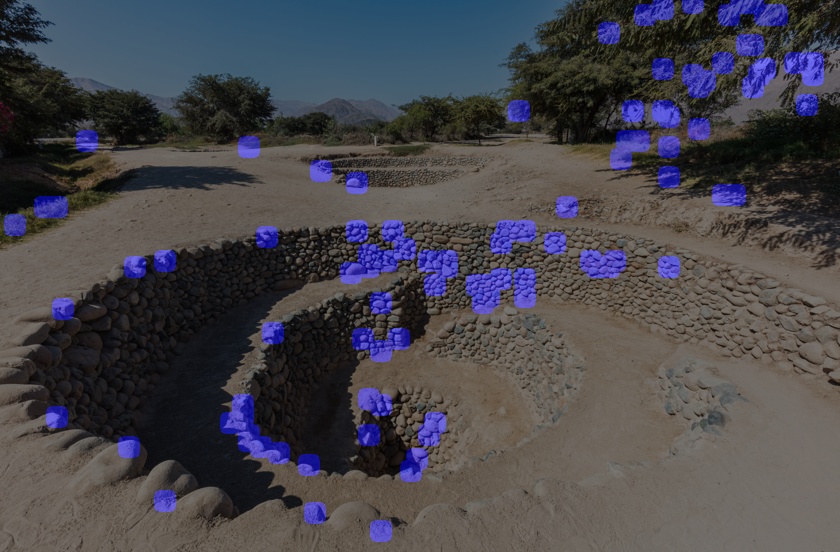}
        \subcaption{Matched result}\label{fig:vis_attn_map}
    \end{subfigure}
    \caption{\textbf{Visualization.} (a) The \roi{} indicated by the red box and the context image as the entire image. (b) The attention map overlaying on the context image where each blue circle represents a strong response with respective to the \roi{} feature.}\label{fig:visualized_attn}
\end{figure}

\begin{figure}
    \centering
    \includegraphics[width=0.6\linewidth]{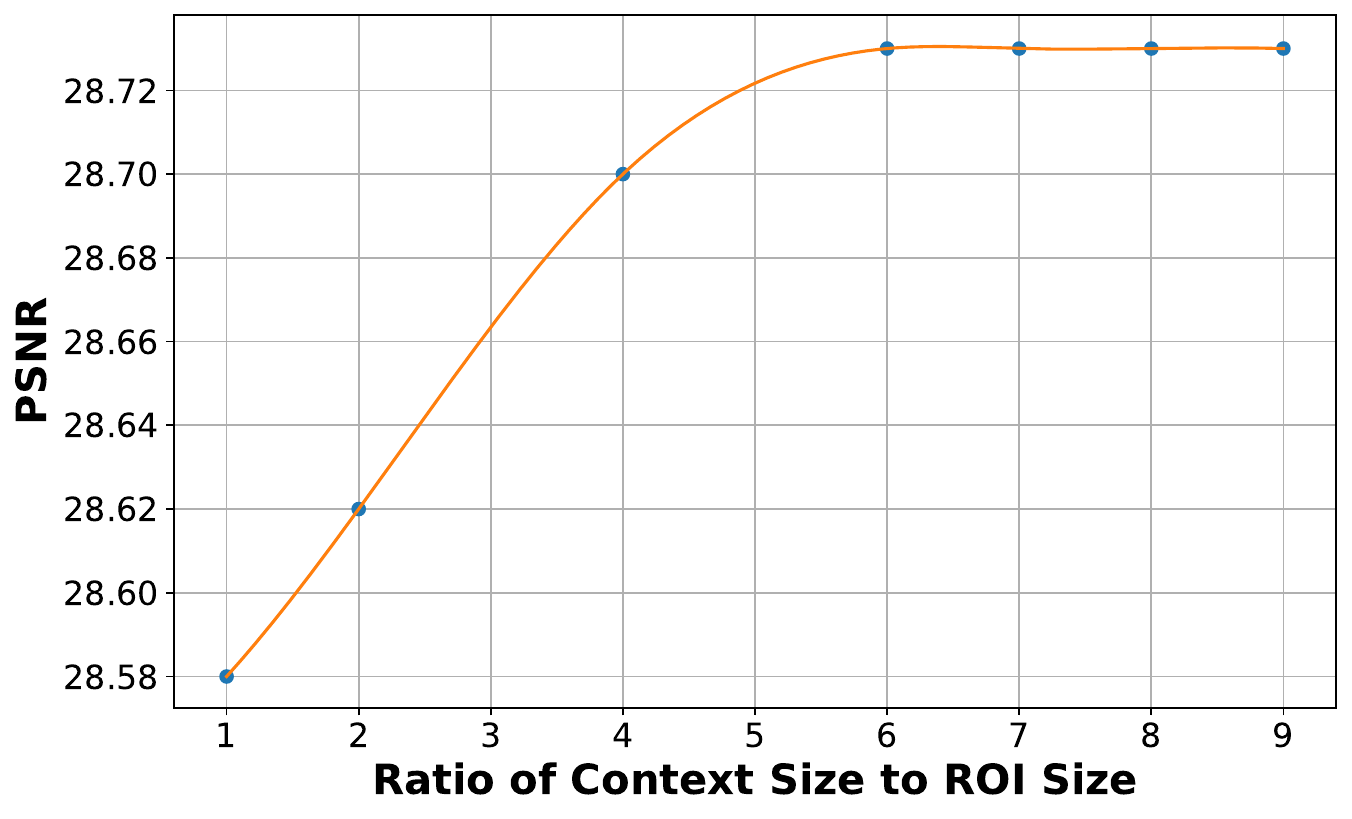}
    \caption{\textbf{Context impact on performance.} We vary the context size and plot the average relationship between the relative context-to-ROI size and PSNR. A saturation point can be observed.}\label{fig:context_impact}
    \vspace{-0.4cm}
\end{figure}

\begin{table}[t!]
  \small
  \centering
  \setlength{\tabcolsep}{5pt}
  \begin{tabular}{c|ccccc}
    \toprule
    $\lambda$  & 0 & 0.1 & 0.5 & 0.8 & 1 \\
    \midrule
    PSNR  & 37.82 & 37.82 & 37.99 & 37.91 & 37.91\\
    SSIM & 0.9927 & 0.9927 & 0.9928 & 0.9929 & 0.9929 \\
    \bottomrule
  \end{tabular}
  \caption{\textbf{Ablation study on Context loss weight $\lambda$.} Both excessively large and small weights detract from the training focus, reducing performance. Optimal results are achieved with $\lambda\!=\!0.5$.} \label{table:context_loss_weight}
  \vspace{-0.5cm}
\end{table}

\subsection{Ablation study}
\textbf{Proposed modules.}
Our ablations are performed on a lightweight SwinIR model, trained from scratch over 200k iterations, to accentuate differences between variants. 
Table~\ref{table:abl_modules} shows the results of using the proposed modules. Without access to the context image, the baseline model is 0.2-0.3dB worse. Combining the two modules achieves the best performance. 
\LB{} boosts the performance by 0.29dB and the \GB{} is 0.19dB, highlighting the more important role of the local information. 
We visualize an example of the attention matching detailed in Section~\ref{sec:global_branch} in Figure~\ref{fig:visualized_attn}. Figure~\ref{fig:vis_input} illustrates the region targeted for restoration, where the entire LR input is used for context, 
while Figure~\ref{fig:vis_attn_map} displays matched results for 
restoring the stone pattern in the \roi{}.
Our matching architecture is relatively simple, yet the majority of the matching outcomes are also stone patterns. 
This observation suggests that the \GB{} gathers relevant information for the restoration process, despite the simplicity of its design. \\

\noindent \textbf{Context impact on local SR.} 
To evaluate the impact of context, we perform experiments by varying the context size. We plot the average relationship between the relative context-to-ROI size and PSNR in Figure~\ref{fig:context_impact}. The results indicate that increasing context size improves performance up to a saturation point -- sepcifically, when the context reaches approximately 6$\times$ the ROI size. Beyond this point, additional context yields diminishing returns. \\

\noindent \textbf{Context loss weight.} Table~\ref{table:context_loss_weight} details an ablation study on the influence of different context loss weights $\lambda$ in Eq~\ref{eq:loss}. Optimal performance is achieved with $\lambda$ values between 0 and 1. Setting $\lambda$ too low, such as at 0 or 0.1, leads to poor results because it obscures the optimization direction for the \LB{}, especially outside the \roi{}. The gradient updates are primarily confined to the \roi{}. 
Conversely, a $\lambda$ set to 1 shifts attention away from the main goal \roi{} restoration. \\

\noindent \textbf{Padding.}
We evaluate the padding using our variant of RCAN on BSD100. 
As shown in Table~\ref{table:comparison_padding_size}, our findings indicate that as the number of surrounding pixels increases, the model's performance improves significantly until it saturates. 
Both the \GB{} and the \LB{} structures contribute to this enhancement, enabling our model to outperform the original architecture by leveraging context information. 
Padding emerges as a consistent and reliable strategy, not only matching but potentially exceeding the performance of models that take LR images as input. This result also highlights that the restoration quality of most image patches may predominantly depend on their surrounding pixels.

\begin{table}[t!]
  \centering
  \setlength{\tabcolsep}{5pt}
  \resizebox{\linewidth}{!}{
  \begin{tabular}{c|ccccc}
    \toprule
    Padding Size & 0 & 2 & 4 & 8 & 12 \\
    \midrule 
    PSNR & 32.22 & 32.36 & 32.40 & 32.42 & 32.43\\
    SSIM & 0.9580 & 0.9585&  0.9586 & 0.9588 & 0.9588\\
    \bottomrule
  \end{tabular}
  }
  \caption{\textbf{Ablation study on padding sizes.} Padding improves the performance up to a saturation point, beyond which further gains are difficult to achieve.} \label{table:comparison_padding_size}
  \vspace{-0.5cm}
\end{table}

\section{Conclusion}
We introduce a specialized task focusing on the super-resolution of a specific local region within a larger image rather than the entire image. We refer to this task as \task{}. This technique is particularly useful when only certain areas need high-resolution enhancement, like surveillance images or photo editing. 
Moreover, \task{} is useful in situations with limited memory resources, allowing for detailed enhancement without processing the whole image.
We address the challenges of \task{} by designing an effective way of aggregating the context. Our approach involves a combination of a base processing branch for the ROI and additional branches that incorporate context information. These branches work together to enhance the \roi{} by utilizing features from the entire image, ensuring efficient and effective restoration. 

Moreover, we observe that tested networks often achieve strong results with padding alone, suggesting limited utilization of global information despite large theoretical receptive fields. We believe enhancing global context utilization could further improve super-resolution performance.

{
    \small
    \bibliographystyle{ieeenat_fullname}
    \bibliography{main}
}


\end{document}